\documentclass{article}
\usepackage{spconf,amsmath,graphicx}
\usepackage{color}
\usepackage{makecell}

\title {Generative Adversarial Network Applications in Creating a Meta-Universe}
\name {Soheyla Amirian\sthanks{Lecturer, Department of Computer Science.}, Thiab R. Taha\sthanks{Professor and Head, Department of Computer Science.}, Khaled Rasheed\sthanks{Director, Institute of Artificial Intelligence; Professor, Department of Computer Science.}, Hamid R. Arabnia\sthanks{Professor Emeritus, Department of Computer Science.}}
\address{\textit{The University of Georgia}\\
Athens, Georgia, USA
}
\begin{document}
%
\maketitle
\begin{abstract}
Generative Adversarial Networks (GANs) are machine learning methods that are used in many important and novel applications. For example, in imaging science, GANs are effectively utilized in generating image datasets, photographs of human faces,
image and video captioning, image-to-image translation, text-to-image translation, video prediction, and 3D object generation to name a few. In this paper, we discuss how GANs can be used to create an artificial world. More specifically, we discuss how GANs help to describe an image utilizing image/video captioning methods and how to translate the image to a new image using image-to-image translation frameworks in a theme we desire. We articulate how GANs impact creating a customized world.
\end{abstract}
\section{Introduction}
\label{sec:intro}
Many deep learning frameworks and architectures are 
utilized by researchers for different applications. Recently, there have been a series of breakthroughs results in various computer vision tasks. Deep learning made an impressive impact on processing images \cite{amirian2018}. 

\begin{figure*}[!ht]
    \centering
    \includegraphics[width=\linewidth]
    {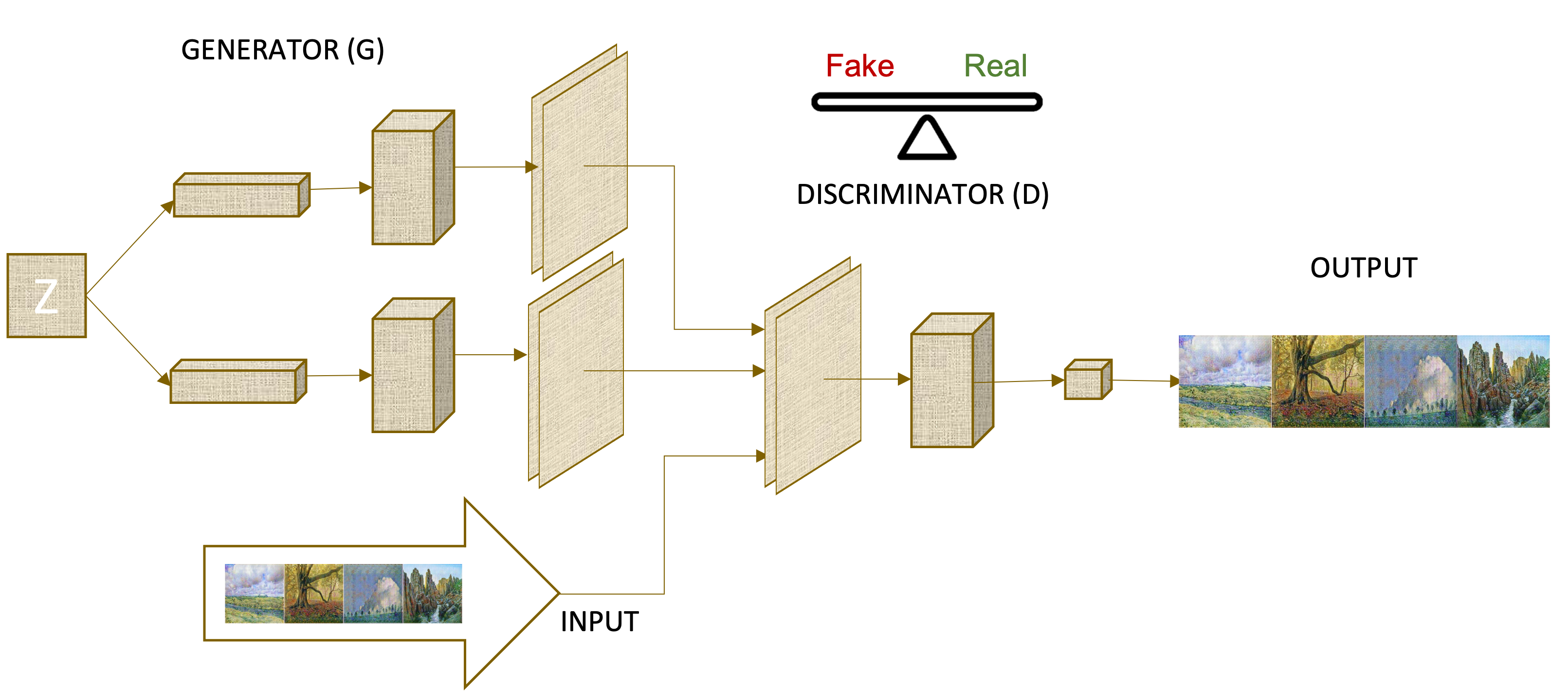}
    \caption{Generative Adversarial Network Architecture. Output: CycleGAN - Monet paintings.}
    \label{fig:GANarch}
\end{figure*}

Generative Adversarial Network is a machine learning model. That was proposed by Goodfellow et al. \cite{goodfellow2014generative} for estimating generative models via an adversarial process for the first time in 2014. 
They simultaneously train two models: a generative model and a discriminative model.
A generative model G captures the data distribution. And a discriminative model D estimates the probability that a sample came from the training data rather than G (See Figure \ref{fig:GANarch}).
Most generative models are trained by adjusting parameters to maximize the probability that the generator net will generate the training data set.
The discriminator is just a regular neural net classifier. The generator takes random noise values $z$ from a prior distribution $P_z$ and maps them to output values $x$ via function $G(z)$. 
Figure \ref{fig:GANarch} illustrates the explanation.

GANs can generate new samples. Most of the applications for GANs have been for images, and the outputs of GANs are imaginary. They are more applicable in many different images problems. They can solve more complex tasks and make them robust. 
Image captioning is the process of generating a concise description of an input picture/image \cite{soh2019image,amirian2019image} as an important task in computer vision. 
Many impactful studies have been done on image captioning \cite{dai2017towards,nezami2019towards}, and video captioning \cite{sung2019adversarial} using GANs. GAN architecture helps to generate more accurate, diverse, and coherent multi-sentence image/video descriptions \cite{soh9281287}. 
A study on developing Equal Opportunity and Fairness in artificial intelligence \cite{toutiaee2020stereotypefree} proposes a novel approach through penalized regression to label stereotype-free GAN-generated synthetic unlabeled images to help to label new data (fictitious output images). They generated their fictitious output images using GAN architecture.

In this research, we introduce recent GAN technologies that are impactful in creating a customized world. The main contribution of this research is the assertion that the Generative Adversarial Network application in image processing is very useful in the Artificial Intelligence (AI) area. Having GAN architecture description, we discuss our previous works in this area. Then, we touch on the recent GAN frameworks applicable to images. We also illustrate different applications of GAN, how it describes images, and how it translates them. Finally, we outline the applications together with advantages of GAN and how they affect the AI world.

%
\section{Discussion}
\label{sec:discussion}
In this section, we demonstrate the use cases of GAN through experiments. We articulate how GAN helps to Generate images. We discuss Generative Adversarial Network models and their applications. We explain different GAN models applied to image captioning and video captioning. Then, we illustrate GAN models in image-to-image Translation. 

\subsection{GAN Application in Generating Images}
GANs have been widely applied to many domains due to their impressive performance especially on the image generation paradigm.
Toutiaee et al. \cite{toutiaee2020stereotypefree} focus on the generative adversarial network (GAN) in their study about "Stereotyping Free Labeling of GAN-made Images" for two reasons; first, it is capable of producing fictitious outputs (imaginary images). Second, it has inspired a legion of scientists to evolve GAN under the impression of producing more realistic-looking data (\cite{karras2019style}). They use the widely used dataset, such as CelebA \cite{liu2015faceattributes}, that is utilized by many GAN practitioners to create supernatural imaginary pictures with 40 face attributes. They experimented super-high-quality fictitious images of humans generated by a state-of-the-art GAN (such as Style-GAN, \cite{karras2019style}). The researchers in Style-GAN tried to improve the quality of output images by proposing an alternative generator architecture in adversarial training. 
\begin{figure}[!ht]
    \centering
    \includegraphics[width=\linewidth]
    {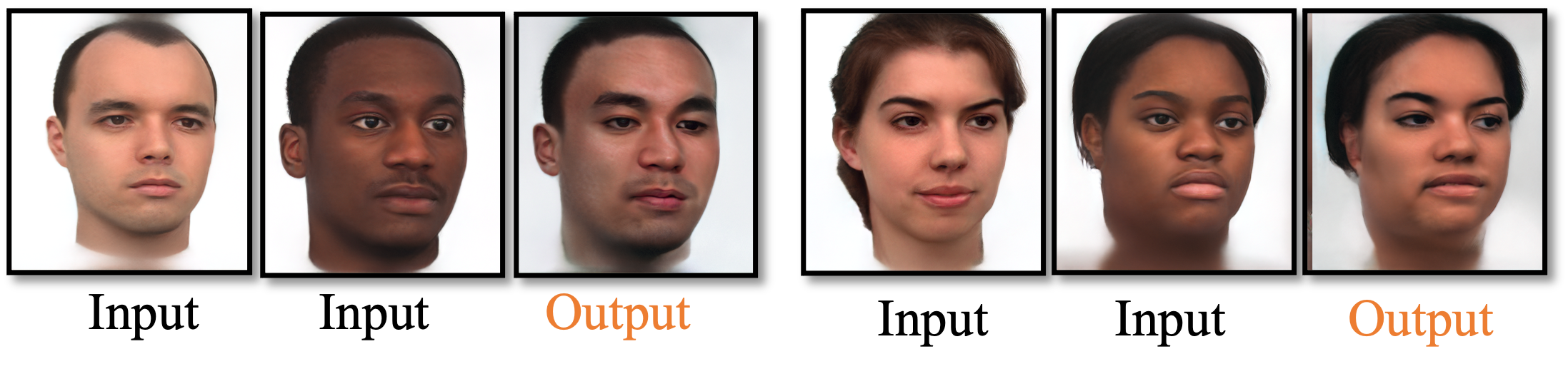}
    \caption{Two examples of fictitious faces generated from adversarial training. By giving people face images, Style-GAN \cite{karras2019style} generates a new imaginary character.}
    \label{fig:StyleGAN}
\end{figure}

Figure \ref{fig:StyleGAN} shows an experiment that has been done by Style-GAN \cite{karras2019style} to illustrate how it helps to create different characters by giving images of people. We trained the model with the Face Place\footnote{http://wiki.cnbc.cmu.edu/Face\_Place} dataset. The results presented in the figure reveal that our produced faces are outstandingly looking natural to viewers.

\begin{figure}[!ht]
    \centering
    \includegraphics[width=\linewidth]
    {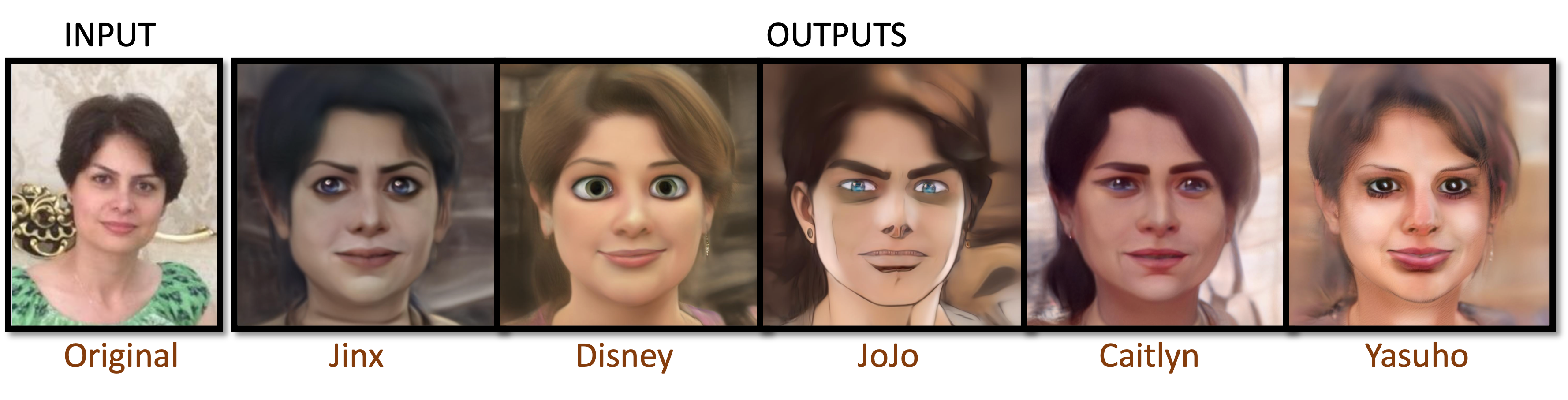}
    \caption{Given a person's face image, JoJoGAN \cite{chong2021jojogan} generates different characters (Jinx, Disney, JoJo, Caitlyn, and Yasuho) using Style-GAN.}
    \label{fig:JoJoGAN}
\end{figure}

Chong et al. \cite{chong2021jojogan} also utilized Style-GAN to propose JoJoGAN. JoJoGAN is a framework for arbitrary one-shot face stylization. Given only a single reference style image, a skilled artist can reproduce new artworks that faithfully capture the style. They train JoJoGAN by approximating a paired training dataset. And then, they finetune a Style-GAN to perform one-shot face stylization. Figure \ref{fig:JoJoGAN} shows an experiment we did by JoJoGAN to show how it helps to create different characters by giving an image of a person.


\subsection{GAN in Image captioning and Video captioning}
Conditional Generative Adversarial Networks (CGAN) presented by Dai et al. \cite{dai2017towards}. This framework jointly learns a generator to produce descriptions conditioned on images and an evaluator to assess how well a description fits the visual content. \cite{dai2017towards} explore an approach to produce sentences that possess three properties: Fidelity in semantics, Naturalness, and Diversity. This work proposed a different task for the GAN method. 
Applying GANs to text generation is nontrivial. It comes with two significant challenges because of the special nature of linguistic representation. First, difficult to apply back-propagation directly that devised via Policy Gradient, originating from reinforcement learning. Second, in the conventional GAN setting, vanishing gradients and error propagation in the training of the generator because of feedbacking from the evaluator which tackled by getting early feedback by an approximated expected future reward through Monte Carlo rollouts. This framework not only results in a generator that can produce natural and diverse expressions but also yields a description evaluator called G-GAN, which is substantially more consistent with human evaluation.
This method is the first in applying the GAN method for image captioning.

In 2019, Nezami et al. \cite{nezami2019towards} propose ATTEND-GAN model. Their contribution is to generate human-like stylistic captions in a two-stage architecture, by ATTEND-GAN using both the designed attention-based caption generator and the adversarial training mechanism on the SentiCap dataset.
The architecture of the ATTEND-GAN model is spatial-visual features that are generated by ResNet-152 network and the caption discriminator is inspired by the Wasserstein GAN (WGAN).
Arjovsky et al. \cite{arjovsky2017wasserstein} claim that the Wasserstein Generative Adversarial Network minimizes a reasonable and efficient approximation of the Earth Mover distance. Also, Wasserstein GAN is able to learn the distribution without mode collapse. Using the WGAN algorithm has significant practical benefits: a meaningful loss metric that correlates with the generator’s convergence and sample quality; improved stability of the optimization process.
Makhzani et al. \cite{makhzani2015adversarial} proposed Adversarial Autoencoder (AAE), which is a probabilistic autoencoder that uses generative adversarial networks to perform variational inference by matching the aggregated posterior of the hidden code vector of the autoencoder with an arbitrary prior distribution, which results in meaningful samples.

By injecting GAN to deep learning, Sung Park et al. \cite{sung2019adversarial} applied Adversarial Networks in their framework by designing a discriminator to evaluate visual relevance to the video, language diversity, fluency, and coherence across sentences. Thus, GAN helps to generate more accurate, diverse, and coherent multi-sentence video descriptions. The task of the discriminator ($D$) is to score the captions generated with the generator ($G$) for a given video. They propose to compose $D$ out of three separate discriminators, each focusing on one of the above tasks. They denote this design as a hybrid discriminator.

\subsection{GAN in image-to-image Translation}

Isola et al. \cite{Isola_2017_CVPR} idea is mapping or translating images to each other. For example, by having a masked image, their approach translates it to a real image or vice versa. One of the applications could be to get a satellite image and translate it to google map. They also show that this approach is effective at synthesizing photos from label maps, reconstructing objects from edge maps, and colorizing images, among other tasks. They use conditional adversarial networks to Pix2Pix image-to-image translation problems.
First, they used Convolutional Neural Networks to minimize Euclidean distance between predicted and ground truth pixels, but it will tend to produce blurry results because Euclidean distance is minimized by averaging all plausible outputs.
For having a good output, like “make the output indistinguishable from reality”, and then automatically learn a loss function appropriate for satisfying this goal, Isola et al. focused on Generative Adversarial Networks. They use conditional adversarial networks to image-to-image translation problems because conditional GANs learn a loss that tries to classify if the output image is real or fake, while simultaneously training a generative model to minimize this loss. The novelty of their work is designing PatchGAN as a discriminator architecture. This discriminator tries to classify if each N × N patch in an image is real or fake.  This is advantageous because a smaller PatchGAN has fewer parameters, runs faster, and can be applied on arbitrarily large images.

CycleGAN is an approach to training image-to-image translation models using the generative adversarial network, or GAN, model architecture. Zhu et al. \cite{Zhu_2017_ICCV} introduced unpaired image-to-image translation using Cycle Consistent Adversarial Networks. They represent an approach for learning to translate an image from a source domain to a target domain in the absence of paired examples. An example application could be using a collection of paintings of a famous artist, learning to render a user’s photograph into their style.
They exploit the property that translation should be
“cycle consistent”, which means if there are translators from X to Y and from Y to X, then these should be inverses of each other, and both mappings should be bijections. They use cycle consistency loss as a way of using transitivity to regularize structured data.
In comparison to their previous work, pix2pix \cite{Isola_2017_CVPR}, which is trained on paired data, shows how close they can get to the “upper bound” without using paired data.
The CycleGAN approach has impressive applications. The CycleGAN can transfer the artistic style from Monet, Van Gogh, Cezanne, and Ukiyo-e to photographs of landscapes. It transforms objects from one class, such as dogs into another class of objects, such as cats, or winter landscapes to summer landscapes. It translates many paintings by Monet to plausible photographs. In addition, it improves the original image in some way.


%

\section{Conclusion and Future Work}
\label{sec:Conclusion}
Generative Adversarial Networks have widely been applied to many domains due to their impressive performance, especially on the image generation paradigm.
GANs can imagine new samples. Therefore, we can get many benefits from them by applying them to a range of applications. These applications could include creating imaginary characters, designing objects, decorating new environments, season translation, object transfiguration, style transfer, and generating photos from paintings \cite{chong2021jojogan,Isola_2017_CVPR,Zhu_2017_ICCV}. 

The main goal of this study was to illustrate how GAN is helpful in virtual reality and augmented reality in the meta-universe we want to create. We presented some experiments. By training the GANs more, we can get more benefits and apply them in different aspects. For example, we can utilize them for people with visual impairments by describing the environments \cite{title2020} using the image and video captioning technologies \cite{soh9281287}.
\section{Acknowledgement}
\label{sec:acknowledgement}
We gratefully acknowledge the support of NVIDIA Corporation with the donation of the Titan V GPU used for this research.

\bibliographystyle{IEEEbib}
\bibliography{refs}

\end{document}